\documentclass[preprint,12pt]{elsarticle}
\usepackage{amsmath,amsfonts, bm}
\usepackage{epstopdf}
\usepackage{url}
\usepackage{graphicx, subfig}
\usepackage{algorithmic}
\usepackage{algorithm}
\usepackage{array}
\usepackage{amssymb}
\usepackage{multirow}
\usepackage{hyperref} 
\usepackage{pifont}
\usepackage{caption}

\usepackage{amssymb}

\journal{Nuclear Physics B}

\begin{document}

\begin{frontmatter}



\title{Multi-Spatio-temporal Fusion Graph Recurrent Network for Traffic forecasting}

\author[mymainaddress,mysecondaryaddress,mythirdaryaddress]{Wei Zhao}
\author[mysecondaryaddress]{Shiqi Zhang\corref{mycorrespondingauthor}}
\cortext[mycorrespondingauthor]{Corresponding author}
\ead{zhangshiqi@gs.zzu.edu.cn}
\author[mymainaddress,mythirdaryaddress]{Bing Zhou}
\author[mysecondaryaddress]{Bei Wang}

\address[mymainaddress]{School of Artificial Intelligence and Computer Science, Zhengzhou University, 450002, China}
\address[mysecondaryaddress]{School of Cyber Science and Engineering, Zhengzhou University, 450002, China}
\address[mythirdaryaddress]{Cooperative Innovation Center of Internet Healthcare, Zhengzhou University, 450002, China}



\begin{abstract}
Traffic forecasting is essential for the traffic construction of smart cities in the new era. However, traffic data's complex spatial and temporal dependencies make traffic forecasting extremely challenging. Most existing traffic forecasting methods rely on the predefined adjacency matrix to model the Spatio-temporal dependencies. Nevertheless, the road traffic state is highly real-time, so the adjacency matrix should change dynamically with time. This article presents a new Multi-Spatio-temporal Fusion Graph Recurrent Network (MSTFGRN\footnote{The code for MSTFGRN is available at \url{https://github.com/zsqZZU/MSTFGRN}.}) to address the issues above. The network proposes a data-driven weighted adjacency matrix generation method to compensate for real-time spatial dependencies not reflected by the predefined adjacency matrix. It also efficiently learns hidden Spatio-temporal dependencies by performing a new two-way Spatio-temporal fusion operation on parallel Spatio-temporal relations at different moments. Finally, global Spatio-temporal dependencies are captured simultaneously by integrating a global attention mechanism into the Spatio-temporal fusion module. Extensive trials on four large-scale, real-world traffic datasets demonstrate that our method achieves state-of-the-art performance compared to alternative baselines.
\end{abstract}


\end{frontmatter}


\section{Introduction}
Rapid urban population expansion has provided a substantial challenge to the urban road traffic infrastructure in light of increased urbanization\cite{ref1}. In the modern era, developing an efficient, intelligent transportation system (ITS\cite{ref2}) has become necessary for constructing smart cities. As an integral part of ITS, traffic forecasting\cite{ref3} has evolved into an active research topic that has the potential to improve the operational efficiency and decision-making of traffic systems\cite{ref4,ref5,ref6}. Nevertheless, complex Spatio-temporal correlations in traffic networks make traffic forecasting a difficult undertaking.

The objective of traffic forecasting is to predict the future state of the road traffic system by analyzing historical traffic state data (e.g., traffic flow, speed, and lane occupancy)\cite{ref7}. Therefore, the early studies considered traffic forecasting a general time series forecasting task. Many traditional research methods have been heavily applied in traffic forecastings, such as History Average model (HA), Vector Auto-Regression (VAR\cite{ref8}), and an Autoregressive Integrated Moving Average (ARIMA\cite{ref9}). All of these methods, however, require the assumption of smoothness\cite{ref10} in time series, which leads to numerous errors in forecasting time series data with large fluctuations and multiple missing values. In recent years, prediction approaches based on deep learning\cite{ref11,ref12} for time series correlation analysis have improved forecast performance. However, they continue disregarding the intricate spatial dependence between road nodes in traffic networks. Researchers\cite{ref7,ref13,ref14} are increasingly turning to integrated models based on Graph Convolutional Networks (GCN\cite{ref15,ref16}) and Recurrent Neural Networks (RNN\cite{ref17}) to describe spatial and temporal dependency, respectively, to capture the Spatio-temporal dependence in traffic flow data. Even though GCN-based traffic forecasting approaches have yielded remarkable results, we believe that two crucial factors are still being neglected.

On the one hand, the GCN-based spatial modeling approach first requires graph convolution operations through predefined adjacency matrices\cite{ref18} to capture spatial dependencies. However, the spatial dependence of the traffic road network is dynamic and largely dependent on the real-time traffic state and the traffic network's topology\cite{ref19}. Therefore, the spatial dependency information of the traffic road network cannot be fully represented by relying solely on the predefined adjacency matrix. On the other hand, the chain structure\cite{ref20} design of RNN and its variant models (e.g., Long Short-Term Memory (LSTM\cite{ref21,ref22}) networks and Gated Recurrent Unit (GRU\cite{ref23})) renders it incapable of learning global features\cite{ref24} and gradient disappearance or gradient explosion may occur when dealing with long-term time series data\cite{ref25,ref26}. In additione, there are typical contextual correlations\cite{ref27,ref28} in traffic events. For instance, the state information of traffic flow will change rapidly if unanticipated events\cite{ref29,ref30} such as traffic accidents and special events. Consequently, the analysis of contextual correlation among traffic data is advantageous for better capturing the Spatio-temporal dependence of traffic data.

To solve the difficulties above, we offer a new Spatio-temporal data forecasting method based on metropolitan road networks for traffic forecasting tasks. Our primary contributions are the following four:

\begin{enumerate}
	\item{We propose a new Semi-autonomous Generation Spatial Adjacency Matrices (SAGSAM). This module autonomously generates weighted adjacency matrices of graphs from real-time traffic data while semi-autonomously generating spatial adjacency matrices for each period in combination with predefined adjacency matrices.}
	\item{We propose a new Semi-autonomous Generative Spatial Graph Convolutional Network (SAGS-GCN). The module overlays multiple layers of GCNs to process the generated spatial adjacency matrix, dynamically capturing the spatial dependencies on each period.}
	\item{We present a Spatio-temporal Fusion Graph Recursive Network (STFGRN). This module replaces the gating unit of GRU with SAGS-GCN to recursively fuse the parallel spatial dependence information on each period to capture the latent local Spatio-temporal dependence.}
	\item{We present a Multi-Spatio-temporal Fusion Graph Recurrent Network (MSTFGRN). This module models the spatial-temporal interdependence of traffic data using a bidirectional STFGRN to identify contextual correlations. In addition, a global temporal attention technique\cite{ref31} is employed to identify global temporal relationships.}
\end{enumerate}

The remaining sections of the paper are organized as follows: In Section 2, we reviewed research and work on traffic prediction challenges. Section 3 discusses the proposed MSTFGRN's structure and implementation in depth. Then, in Section 4, we conduct extensive comparison tests between MSTFGRN and various baseline models across multiple datasets, including ablation studies and parameter investigations. The paper is finally summarized in Section 5.

\section{Related Works}
Due to the complex geographical correlation between traffic road networks, traffic forecasting approaches that merely consider temporal correlation have failed to estimate road traffic conditions on roadways effectively. In recent years, Convolutional Neural Networks (CNN)\cite{ref32} and Graph Neural Networks (GNN)\cite{ref16} has been used to capture spatial dependence and has achieved good results in many works\cite{ref7,ref13,ref33}. Based on the above analysis, we will present some working methods for related traffic forecasting in terms of both time and Spatio-temporal dependence.

$\mathbf{Temporal\ dependency\ modeling}$: Most such studies\cite{ref21,ref22,ref23} rely on the recursive structure of LSTM or GRU to sequentially process traffic sequence data to capture the time dependence. M-B-LSTM\cite{ref28} mitigates the overfitting and gradient disappearance and gradient explosion problems that occur in traditional recurrent networks during feature learning by constructing online self-learning networks and introducing bidirectional long-short memory networks. Unlike recurrent neural networks, some works use temporal convolutional networks (TCN\cite{ref34,ref35}), allowing the models to utilize less time to process longer sequential information. In recent works\cite{ref36,ref37}, the introduction of Transformer-based\cite{ref31} time series prediction models can effectively capture the long-term time dependence between the output and input of long time series.

$\mathbf{Spatiotemporal\ dependency\ modeling}$: Modeling traffic data's spatial and temporal dependence is central to traffic forecasting. In some works\cite{ref33,ref38}, traffic road networks are described as two-dimensional grids, and CNNs are used to model the spatial dependence of the two-dimensional grid regions. However, the partitioning of topologically structured traffic road networks using two-dimensional grids may cause the problem of edge feature loss. In order to develop traffic forecasting methods generalized to graph topologies, more and more researchers have turned to investigating GCN-based Spatio-temporal prediction methods in recent years. Among them, DCRNN\cite{ref14} captures spatial and temporal dependencies in traffic flow data utilizing GCN and GRU. Following DCRNN, ASTGCN\cite{ref13} and STSGCN\cite{ref39} further add Spatio-temporal attention mechanisms to capture dynamic Spatio-temporal dependencies. However, they both model the spatial dependencies of traffic data using predefined adjacency matrices. Therefore Graph Wavenet\cite{ref40} and AGCRN\cite{ref33} use an adaptive adjacency matrix that can capture spatial dependencies without a predefined adjacency matrix. In addition, STFGNN\cite{ref41} proposes a data-driven method for generating temporal graphs to compensate for spatial dependency information that a predefined adjacency matrix may not reflect. In recent works, STGODE\cite{ref42} measures the semantic similarity of each time series using Dynamic Time Warping (DTW\cite{ref43}), which is then used as the weight in the semantic adjacency matrix to identify more significant spatiotemporal relationships. z-GCNETs\cite{ref44} incorporate the concepts of time-aware zigzag persistence into time-aware GCN and produce an excellent performance on traffic forecasting.

In contrast to the above work, our proposed model can dynamically generate spatial adjacency matrices for corresponding moments based on real-time traffic state information on individual roads, and accurately capture the Spatio-temporal dependence of the captured traffic data through a new Spatio-temporal fusion network.
\section{Methods}
\subsection{Problem Definition}
$\mathbf{Definition\ 1}$: The information describing the topological structure of the traffic road network is represented in graph \(G=(V, E, A)\). The set \(V=\{v_1,v_2,\cdots,v_N\}\) represents all road nodes on the road network topology graph, whereas \(N\) denotes the total quantity of road nodes. \(E\) represents the set of edge-to-node connection relationships. \(A \in R^{N \times N}\) denotes the adjacency matrix of graph \(G\), which contains only two numbers, \(0\) and \(1\). That is, for any road node \(v_{i}\) and \(v_{j}\), when two nodes are connected: \(A(i,j)=A(j, i)=1\), and vice versa is \(0\).

$\mathbf{Definition\ 2}$: Using the traffic information on the traffic routes as the feature characteristics in the network nodes, the feature matrix \(X \in R^{N \times T \times 1}\) is built, where \(T\) represents the length of the historical time series, \(X_t \in R^{N \times 1}\) represents the collection of traffic speed statistics at the time \(t\) for \(N\) road nodes in the traffic network. 

With the formulation above, the traffic prediction problem can be understood using the traffic network topology graph \(G\) and the feature matrix \(X\) to anticipate the traffic state information for the following \(T\) time steps via the mapping function \(f(\cdot)\).

\begin{equation}
	\label{deqn_ex1}
	[X_{t+1},\cdots,X_{t+T}]=f (G;(X_{t-T},X_{t-(T-1)},\cdots,X_{t}))
\end{equation}

\subsection{The Model Architecture}
\begin{figure}[H]
	\centering
	\includegraphics[width=5.2in]{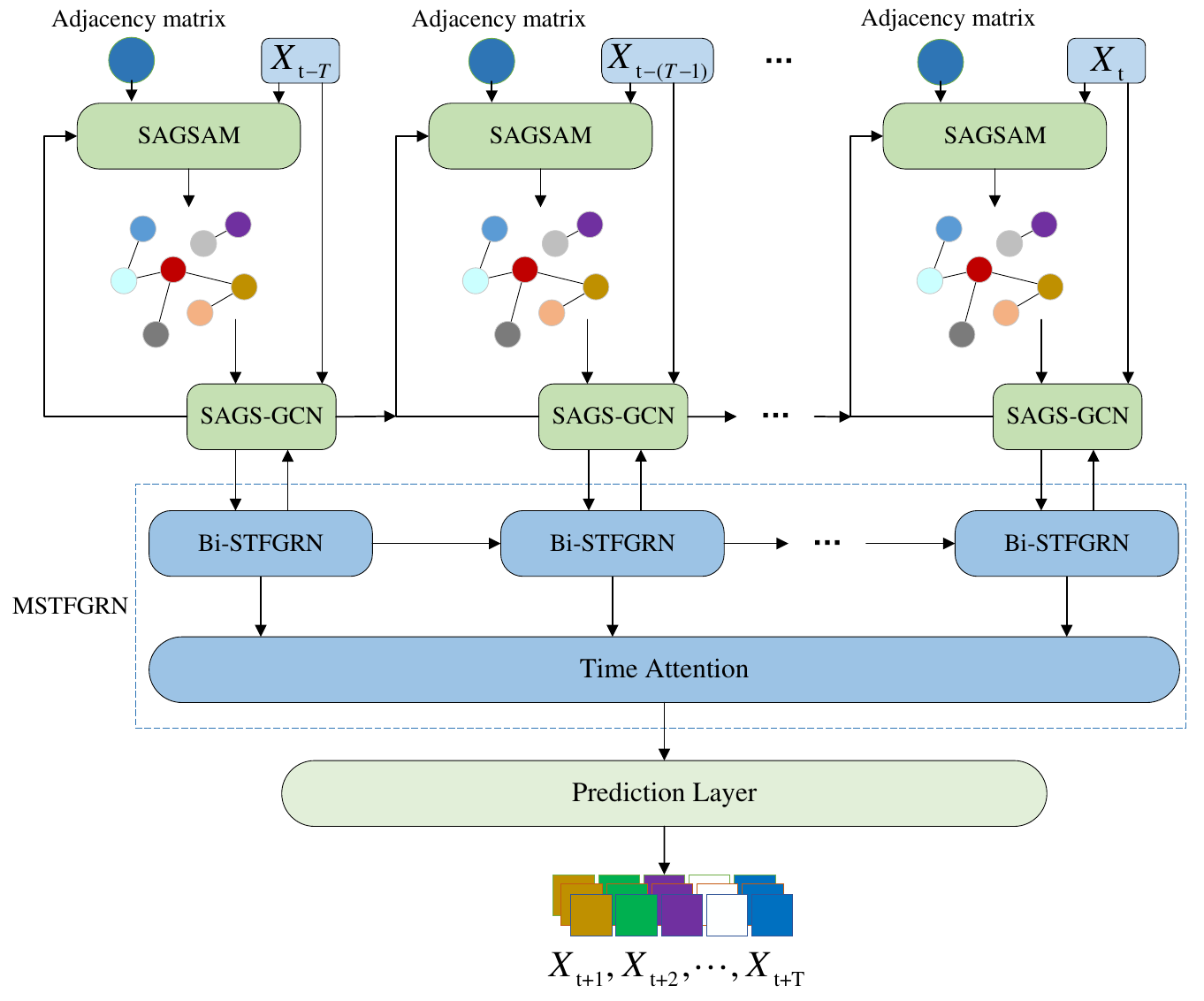}
	\caption{The Proposed Multi-Spatio-temporal Fusion Graph Recurrent Network Framework.}
	\label{fig_1}
\end{figure}

This unit will describe how to implement the traffic forecasting task using the MSTFGRN. As shown in Figure \ref{fig_1}, the model has three primary components: 1) SAGS-GCN layer: seeks to capture the spatial dependence between traffic road network nodes at each time; 2) MSTFGRN layer: its objective is to comprehensively capture the Spatio-temporal dependence of traffic data by fusing parallel Spatio-temporal relationships at each instant; 3) Prediction layer: a neural network with full connectivity is employed to output prediction results.
\subsubsection{Spatial Dependency Modeling}
\begin{figure}[H]
	\centering
	\includegraphics[width=3.0in]{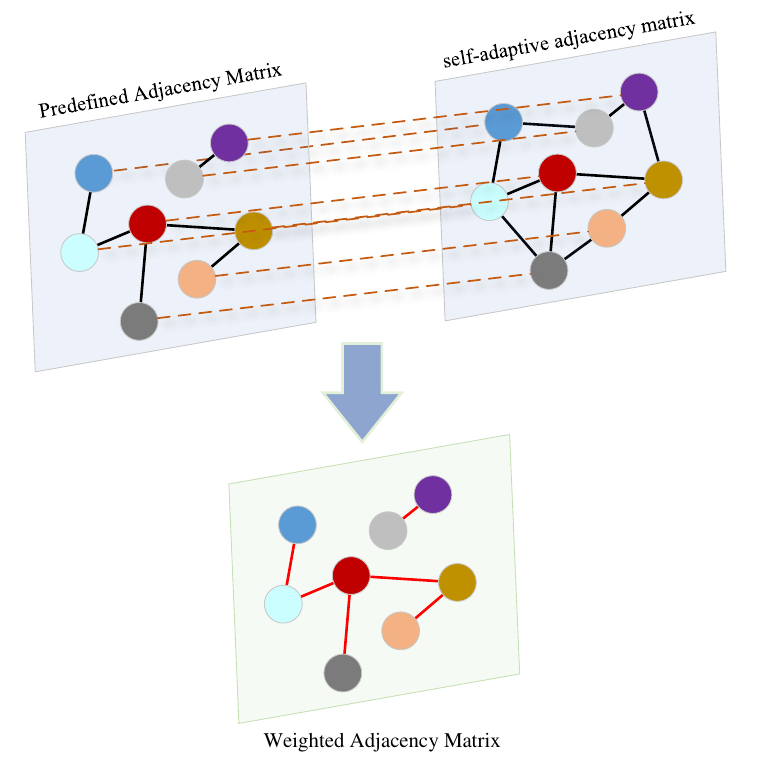}
	\caption{Semi-automatically Generated Spatial Adjacency Matrix.}
	\label{fig_2}
\end{figure}
As illustrated in Figure \ref{fig_2}, we propose a Semi-autonomous Generation Spatial Adjacency Matrix (SAGSAM) to effectively capture the spatial relationships of individual time steps. SAGSAM can autonomously generate the weighted adjacency matrix for the corresponding moment based on the traffic state information on the road at different moments and integrate it with the predefined adjacency matrix to dynamically generate the weighted adjacency matrix specific to different moments. The specific calculation process is shown in Eq. \ref{deqn_ex2}.

\begin{equation}
	\label{deqn_ex2}
	\tilde{A}=softmax(ReLU(E_A \cdot E_A^T))\cdot A
\end{equation}
Where \(E_A \in R^{N \times d}\) indicates that all nodes can learn the Embedded dictionary, \(d\) is the node embedding dimension. \(softmax\) is the normalized exponential function, and \(ReLU\) is the nonlinear activation function. \(E_A^T\) denotes the transpose matrix of \(E_A\). \(\tilde{A} \in R^{N \times N}\) is the generated spatial adjacency matrix.

Meanwhile, we further propose the Semi-autonomous Generative Spatial Graph Convolution Network (SAGS-GCN) by combining the SAGSAM module with NAPL-GCN\cite{ref33} to capture the spatial dependence at each parallel time step. Specifically, the weighted adjacency matrix \(\tilde{A}_{t}\) generated by SAGSAM at any time step \(t\) and the feature information \(X_t \in R^{N \times C}\) of all nodes at the corresponding moment are used as the input information of SAGS-GCN, and the output \(Z_t \in R^{N \times F}\) is computed by Eq. \ref{deqn_ex3}.

\begin{equation}
	\label{deqn_ex3}
	Z_t=(I_N + softmax(ReLU(E_{At} \cdot E_{At}^T))\cdot A)X_tE_{At}\cdot W_{At} +E_{At}\cdot  b_{At}
\end{equation}
Where \(C\) denotes the input feature dimension of each node and \(F\) is the final output feature dimension after graph convolution operation. \(E_{At}\) is the node embedding matrix. \(I_N \in R^{N \times N}\) is the diagonal matrix. \(W_{At} \in R^{d \times C \times F}\) is the shared weight pool, and the weight parameter \(W_t \in R^{N\times C \times F}\) of all nodes can be obtained by \(E_{At}\cdot W_{At}\). correspondingly, \(b_{At} \in R^{d \times F}\) is the shared offset term pool, and the offset term \(b_t \in R^{C \times F}\) of all nodes can be obtained by \(E_{At}\cdot b_{At}\).

\subsubsection{Multi-Spatio-temporal Dependency Modeling}
\begin{figure}[H]
	\centering
	\includegraphics[width=4.0in]{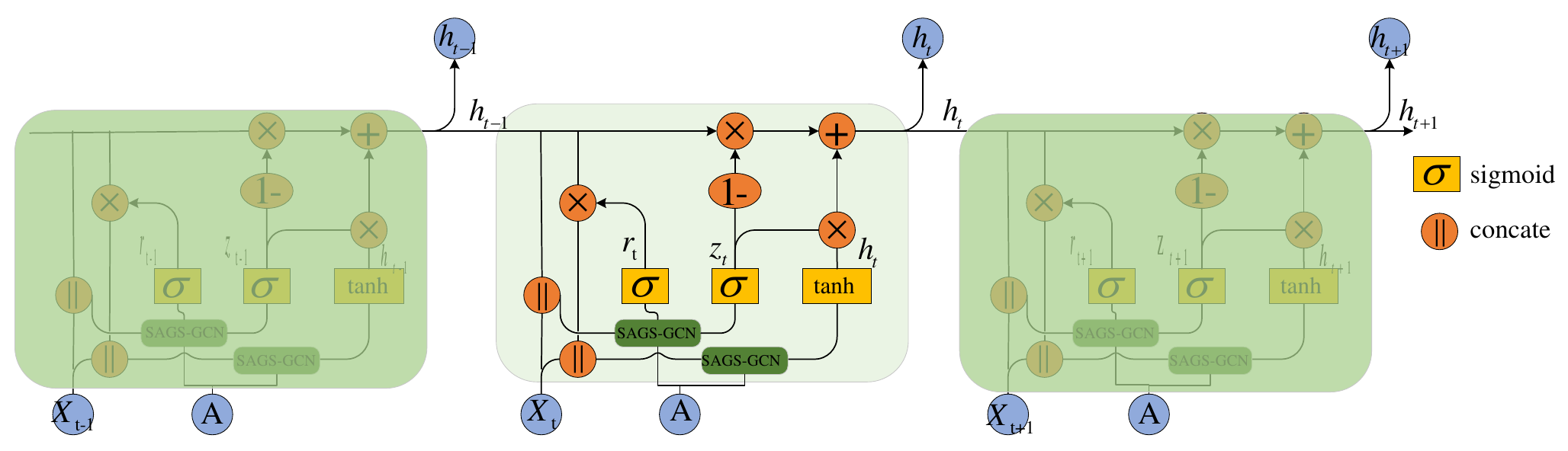}
	\caption{Spatiotemporal Fusion Graph Recurrent Network.}
	\label{fig_3}
\end{figure}
To simultaneously capture traffic data's Spatio-temporal dependencies simultaneously, we propose a Spatio-temporal Fusion Graph Recursive Network (STFGRN). As shown in Figure \ref{fig_3}, this network replaces the gating unit of GRU with SAGS-GCN. The STFGRN recursively performs a new fusion operation on the parallel Spatio-temporal dependencies at each adjacent time step to efficiently learn the hidden local Spatio-temporal dependencies in the traffic data. Meanwhile, to capture contextual Spatio-temporal correlations between traffic events, we model the Spatio-temporal dependencies of traffic networks using forward and reverse STFGRN. To facilitate the description, we introduce the computational process of STFGRN using the forward STFGRN as an example. Specifically, at any moment \(t\), the given \(X_t\) at the current time and the output \(\overrightarrow{h}_{t-1} \in R^{N \times F^\prime}\) of STFGRN at the previous moment is concatenated as the input information of the current moment, and then the following calculation is performed.  

\begin{equation} \label{deqn_ex4}
	\begin{split}
		\tilde{A}&=I_N + softmax(ReLU(E_{At} \cdot E_{At}^T))\cdot A \\ 
		z_t&=\sigma(\tilde{A}[X_t, \overrightarrow{h}_{t-1}]E_{At}\cdot W_{zt} +E_{At}\cdot  b_{zt}) \\
		r_t&=\sigma(\tilde{A}[X_t, \overrightarrow{h}_{t-1}]E_{At}\cdot W_{rt} +E_{At}\cdot  b_{rt}) \\
		\widetilde{h_t}&=tanh(\tilde{A}[X_t, r_t \odot \overrightarrow{h}_{t-1}]E_{At}\cdot W_{\widetilde{h}t} +E_{At}\cdot  b_{\widetilde{h}t}) \\
		\overrightarrow{h_t}&=z_t\odot \overrightarrow{h}_{t-1} + (1-z_t)\odot \widetilde{h_t} 
	\end{split}
\end{equation}
Where \(\sigma(\cdot)\) is the sigmoid activation function, \(E_{At}\), \(W_{zt}\), \(W_{rt}\), \(W_{\widetilde{h}t}\), \(b_{zt}\), \(b_{rt}\), and \(b_{\widetilde{h}t}\) are learnable parameters. \(\widetilde{h_t}\) is the candidate's hidden layer state. $\overrightarrow{h_t} \in R^{N \times F^\prime}$ is the output at the current moment. $[\cdot]$ denotes the concate operation in the feature dimension, and $\odot$ denotes the multiplication by elements. Waiting for the last step to complete the operation, we get the forward output result $\overrightarrow{H} = [\overrightarrow{h}_{t-T},\overrightarrow{h}_{t-(T-1)},\cdots,\overrightarrow{h}_{t}]$, $\overrightarrow{H} \in R^{N\times T \times F^\prime}$. The reverse operation is the same as the forward operation, and the forward and reverse outputs are concatenated by Eq. \ref{deqn_ex5} to obtain the output result $H \in R^{N \times T\times 2F^\prime}$ of the network finally.

\begin{equation}
	\label{deqn_ex5}
	H = \overrightarrow{H} \ || \ \overleftarrow{H}
\end{equation}

Besides, to simultaneously capture the global Spatio-temporal dependencies, we add a global attention mechanism after Bi-STFGRN to form the proposed Multi-Spatio-temporal Fusion Graph Recurrent Network (MSTFGRN). Specifically, after capturing the local Spatio-temporal dependence, we impose a self-attentive mechanism\cite{ref45} on each node and then aggregate it with the original information to capture the global Spatio-temporal dependence by processing the entire sequence data on the nodes in parallel. For any node $v_i$, the specific calculation procedure is shown in Eq. \ref{deqn_ex6}.
\begin{equation} \label{deqn_ex6}
	\begin{split}
		Q_{v_i} &= H_{v_i,:,:}W_q+b_q \\
		K_{v_i} &= H_{v_i,:,:}W_k+b_k \\
		V_{v_i} &= H_{v_i,:,:}W_v+b_v \\
		A_{v_i,:,:} &= softmax(\frac{Q_{v_i}K^{T}_{v_i}}{\sqrt{d}})V_{v_i}\\
		\widetilde{H}_{v_i,:,:} &=normLayer(A_{v_i,:,:}  +H_{v_i,:,:}) 
	\end{split}
\end{equation}
where $H_{v_i,:,:} \in R^{T \times 2F^\prime}$ is the full time series information of the node, \(Q_{v_i}\), \(K_{v_i}\), and \(V_{v_i}\) are querry, key, and value, respectively, \(W_q\), \(W_k\), \(W_v\), \(b_q\), \(b_k\) and \(b_v\) are learnable parameters, and \(\sqrt{d}\) is the scale set. $\widetilde{H}_{v_i,:,:} \in R^{T \times 2F^\prime}$ is the output information of the node, and $normLayer$ is the normalization operation. Until all nodes finish the computation, we get the final output $\widetilde{H} \in R^{N \times T \times 2F^\prime}$.

\subsection{Multi-step Traffic Forecasting}
Finally, we implement the multi-step traffic prediction task by performing linear variations of $\widetilde{H} \in R^{N \times T \times 2F^\prime}$ by a fully connected neural network.

\begin{equation}
	\label{deqn_ex7}
	Y^\prime =W_f \cdot \widetilde{H} +b_f
\end{equation}
Where \(W_f\) and \(b_f\) are the weight matrix and bias terms, and $Y^\prime \in R^{N \times T \times 1}$ is the final prediction result.

The objective of the training is to narrow the gap between the actual road traffic speed, \(Y\), and the expected value, \(Y^\prime\). In this paper, the \(L1\) loss function is chosen for backpropagation-based neural network model optimization. The specific calculating method is illustrated in Eq. \ref{deqn_ex8}.

\begin{equation}
	\label{deqn_ex8}
	loss =\frac{1}{T}\sum_{i=1}^{T}|Y_{t+i} - Y_{t+i}^\prime|
\end{equation}

\section{Experiment}
\subsection{Datasets}
To prove the efficacy of the proposed framework, we conducted experiments using PeMS03, PeMS04, PeMS07, and PeMS08, four public traffic network datasets. All traffic flow information is received from the Caltrans Performance Measurement System (PeMS), which compiles data gathered every five minutes by roadway sensors. Table \ref{tab1} summarizes some critical statistics for these four datasets.
\begin{table}[htb]
	\begin{center}
		\caption{Datasets statistics}
		\label{tab1}
		\begin{tabular}{| c | c | c |c | c |}
			\hline
			Datasets & Sensors&Edges&Unit&Time Steps\\
			\hline
			PeMS03&358&547&5 min&26208\\
			
			\hline
			PeMS04&307&340&5 min&16992\\
			\hline
			PeMS07&883&866&5 min&28224\\
			
			\hline
			PeMS08&170&295&5 min&17856\\
			\hline
		\end{tabular}
	\end{center}
\end{table}
\subsection{Baseline Methods}
MSTFGRN was compared to some of the most advanced baseline models. The following is a summary of these baselines.
\begin{itemize}
	\item
	HA: The predicted outcome used by the model is the average of past traffic data.
	\item
	VAR\cite{ref8}: The model is a standard time series model that captures the pairwise temporal dependence between time series.
	\item
	FC-LSTM\cite{ref21}: An LSTM with a fully connected layer is used to accomplish the traffic prediction task.
	\item
	TCN\cite{ref34}: The model uses inflated convolution to obtain a larger perceptual field with less cost.
	\item
	DCRNN\cite{ref14}: The model incorporates spatial correlation through bidirectional random wandering on the graph and temporal correlation through GUR.
	\item
	ASTGCN\cite{ref13}: The model further introduces spatial and temporal attention mechanisms to dynamically model spatial and temporal dependencies.
	\item
	STSGCN\cite{ref39}: The model uses local Spatio-temporal subgraph modules to model local correlations independently.
	\item
	AGCRN\cite{ref33}: The model captures the fine-grained Spatio-temporal correlation of specific nodes in a traffic sequence.
	\item
	STFGNN\cite{ref41}: The approach provides a temporal graph based on the similarity of time series and includes a CNN with gated expansion to capture local and global relationships.
	\item
	STGODE\cite{ref42}: The model proposes a continuous representation that increases the GCN's depth, expanding the GCN's perceptual field to capture deeper Spatio-temporal dependencies.
	\item
	Z-GCNETs\cite{ref44}: The model develops a zigzag topology layer for time-aware graphical convolutional networks to capture the complex Spatio-temporal dependencies.
\end{itemize}

\subsection{Experimental Settings}
We normalize all datasets with Z-score and divide each dataset into training sets of  $60\% $, validation sets of $20\% $, and test sets of $20\% $. Then, a \(T+T\) window is slid over the separated datasets (\(T\) consecutive time steps are utilized to predict the traffic situation information for the subsequent \(T\) consecutive time steps.). Here, we set the size of \(T\) to 12. The embedding dimension of our model's nodes is set to $10$, and the size of all hidden layers is set to $64$. The batch size is set to $64$, the learning rate is set to $0.001$, and the Adam optimizer is used to optimize the model with a maximum of $100$ iterations.

All comparison experiments are configured according to their open-source code and optimal hyperparameter values in theory, all on a server with Ubuntu 18.04.6 with an Intel Core i5-10500 @ 3.10GHz CPU and NVIDIA GeForce 2080Ti GPU 11GB. In addition, we use the following three metrics to measure the model's predictive performance.

\begin{itemize}
	\item
	Mean Absolute Error(MAE):
\end{itemize}
\begin{equation}
	\label{deqn_ex13}
	MAE=\frac{1}{K}{\overset{K}{\underset {i=1}{\sum}}}|Y_i-Y^\prime_i|
\end{equation}
\begin{itemize}
	\item
	Root Mean Squared Error(RMSE):
\end{itemize}
\begin{equation}
	\label{deqn_ex14}
	RMSE=\sqrt{\frac{1}{K} \sum^K_{i=1}(Y_i-Y^\prime_i)}
\end{equation}

\begin{itemize}
	\item
	Mean Absolute Percentage Error(MAPE):
\end{itemize}
\begin{equation}
	\label{deqn_ex15}
	MAPE=\frac{100 \%}{K} \overset{K}{\underset {i=1}{\sum}} |\frac{Y_i-Y^\prime_i}{Y_i}|
\end{equation}
Where \(K\) denotes the total number of samples. The lower the values of the three indicators above, the greater the model's predictive accuracy. We conduct each experiment five times and then calculate the mean value as the test result.

\subsection{Experiment Results and Analysis}
\begin{table*}[htb]
	\begin{center}
		\renewcommand{\arraystretch}{1.6}
		\caption{Comparison of MSTFGRN and baseline models' performance on the PeMS03, PeMS04, PeMS07, and PeMS08 datasets}
		\label{tab2}
		\resizebox{\textwidth}{!}{
			\begin{tabular}{cccccccccccccc}
				\hline
				\multirow{2}{*}{Model} & Dataset & \multicolumn{3}{c}{PeMS03}  & \multicolumn{3}{c}{PeMS04}    & \multicolumn{3}{c}{PeMS07}  & \multicolumn{3}{c}{PeMS08}   \\ \cline{2-14}
				
				& Metrics & MAE      & RMSE     & MAPE     & MAE      & RMSE     & MAPE & MAE      & RMSE     & MAPE     & MAE      & RMSE     & MAPE    \\
				\hline
				\multicolumn{2}{c}{HA}           & 31.74    & 51.79    & 33.49\%  & 39.87    & 59.04    & 27.59\% & 45.32    & 65.74    & 23.92\% & 35.16    & 59.74    & 28.35\%  \\
				\hline
				\multicolumn{2}{c}{VAR}          & 23.75    & 37.97    & 24.53\%  & 24.61    & 38.61    & 17.54\% & 49.89   & 75.45    & 32.13\% & 19.21    & 29.84    & 13.13\% \\
				\hline
				\multicolumn{2}{c}{FC-LSTM}          & 20.96    & 36.01    & 20.76\%  & 25.01    & 41.42    & 16.18\%  & 33.26    & 59.92    & 14.32\% & 23.49    & 38.89    & 14.55\% \\
				\hline
				\multicolumn{2}{c}{TCN}          & 19.32    & 33.55    & 19.93\%  & 23.22    & 37.26    & 15.59\% & 32.27    & 42.23    & 14.26\% & 22.72    & 35.79    & 14.03\% \\
				\hline
				\multicolumn{2}{c}{DCRNN}      & 17.48    & 29.19    & 16.83\%  & 21.22    & 33.44    & 14.17\% & 24.69    & 37.88    & 10.80\% & 16.82    & 26.32    & 10.92\% \\
				\hline
				\multicolumn{2}{c}{ASTGCN}          & 17.65    & 29.63    & 16.94\%  & 22.03    & 34.99    & 14.59\% & 24.01    & 37.87    & 10.73\% & 18.36    & 28.31    & 11.25\% \\
				\hline
				\multicolumn{2}{c}{STSGCN}        & 17.48    & 29.21    & 16.78\%  & 21.19    & 33.65    & 13.90\% & 24.26    & 39.03    & 10.21\% & 17.13    & 26.80    & 10.96\% \\
				\hline
				\multicolumn{2}{c}{AGCRN}        & \underline{15.97}    & 28.11    & \underline{15.23\%}   & 19.83    & 32.26    & 12.97\% & \underline{21.13}    & 35.20    & \underline{8.96\%} & 15.95    & 25.22    & 10.09\% \\
				\hline
				\multicolumn{2}{c}{STFGNN}       & 16.77    & 28.34    & 16.30\%  & 20.18    & 32.41    & 13.94\%  & 22.07    & 35.80    & 9.21\% & 16.64    & 26.25    & 10.60\% \\
				\hline
				\multicolumn{2}{c}{STGODE}        & 16.32    & \underline{27.23}    & 16.25\%  & 20.95    & 32.66    & 14.95\% & 22.90    & 37.54    & 10.14\% & 16.81    & 25.97    & 10.62\% \\
				\hline
				\multicolumn{2}{c}{Z-GCNETs}       & 16.64    & 28.15    & 16.39\%  & \underline{19.50}    & \underline{31.61}    & \underline{12.78\%} & 21.77    & \underline{35.17}    & 9.25\%& \underline{15.76}    & \underline{25.11}    & \underline{10.01\%}  \\
				\hline
				\multicolumn{2}{c}{$\mathbf{MSTFGRN}$}       & $\mathbf{15.10}$    & $\mathbf{26.54}$    & $\mathbf{14.07\%}$  & $\mathbf{19.03}$    & $\mathbf{30.95}$    & $\mathbf{12.76\%}$ & $\mathbf{20.34 }$   & $\mathbf{34.29}$   & $\mathbf{8.53\%}$& $\mathbf{15.43 }$   & $\mathbf{24.77}$    & $\mathbf{9.97\%}$  \\
				\hline
				
		\end{tabular}}
	\end{center}
\end{table*}
Table \ref{tab2} displays the performance metrics of MSTFGRN compared to 11 other models for 12-time step (60-minute) predictions on the PeMS03, PeMS04, PeMS07, and PeMS08 datasets. The results show that our proposed MSTFGRN achieves optimal results compared to each of the baseline models, demonstrating the feasibility of MSTFGRN as a novel Spatio-temporal prediction model for traffic forecasting tasks. At the same time, the analysis of our experimental findings permits us to notice the following phenomena:

\begin{itemize}
	\begin{figure}[H]
		\centering
		\includegraphics[width=5.5in]{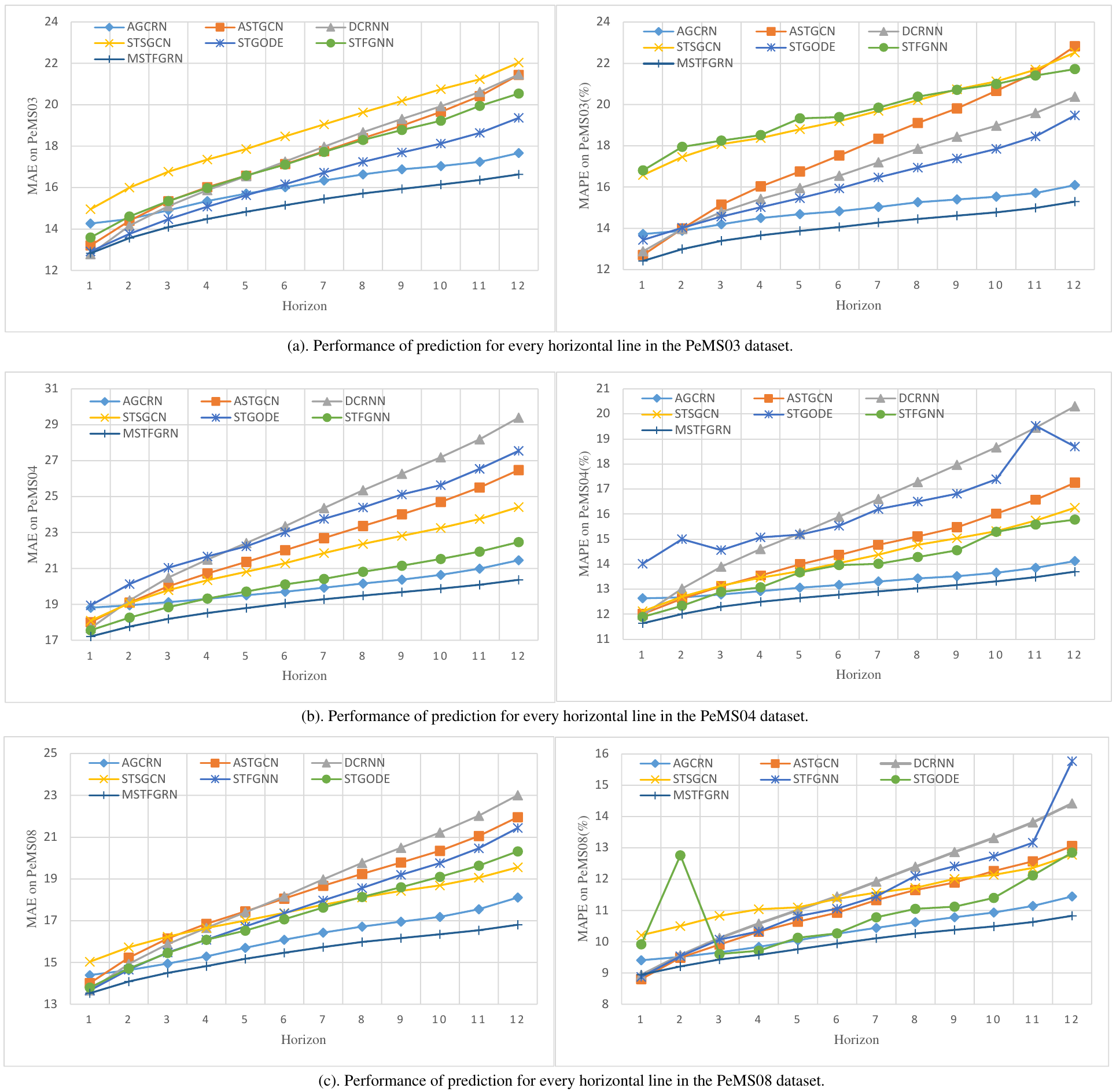}
		\caption{Metrics on PeMS03, PeMS04, and PeMS08 dataset.}
		\label{fig_4}
	\end{figure}
	\item [1)] 
	$\mathbf{Spatiotemporal \ modeling \ capabilities}$: According to Table \ref{tab2}, models based on deep learning typically have higher prediction accuracy than statistical models (such as HA and VAR models). In addition, the prediction accuracy of Spatio-temporal correlation modeling-based approaches (such as DCRNN, ASTGCN, and STSGCN) is much higher than that of temporal correlation modeling-based methods (LSTM and TCN). Similarly, MSTFGRN achieves the best prediction performance across all assessment measures among all Spatio-temporal prediction models. Among them, MSTFGRN has a $5.45 \% $ year-on-year reduction in MAE, $2.53 \% $ year-on-year reduction in RMSE, and $7.62 \% $ year-on-year reduction in MAPE on PeMS03 compared to the best results in other baseline models. On the PeMS07 dataset, which has the highest number of nodes and the most complex data, MSTFGRN shows a $3.75 \% $ reduction in MAE, $2.50 \% $ reduction in RMSE, and $4.80 \% $ reduction in MAPE compared to the optimal results of other models. This proves that the MSTFGRN model has better Spatio-temporal modeling capability than other advanced Spatio-temporal prediction models. 
	\item [2)]
	$\mathbf{Long-term \ forecasting \ ability}$: As shown in Figure \ref{fig_4}, which compares the prediction performance of MSTFGRN to that of other Spatio-temporal prediction models on various Horizons, the three performance curves of MSTFGRN exhibit relatively tiny oscillation trends on each data set, showing that our proposed technique is insensitive to the prediction horizons and the prediction performance is relatively stable. This permits the MSTFGRN model to be utilized for short-term and long-term forecasting. To verify the long-term prediction ability of MSTFGRN, we selected the AGCRN for comparison. We visualized the predicted output of MSTFGRN and AGCRN for 288 consecutive time steps (i.e., 24 hours) with the actual values at any of the same nodes. From Figure \ref{fig_5}, we can observe that MSTFGRN usually fits the actual values better than AGCRN and learns the traffic flow data variation pattern relatively well when the actual value curve fluctuates more drastically. This proves that the MSTFGRN model has good Spatio-temporal modeling capability in long-time forecasting tasks.
\begin{figure}[H]
	\centering
	\includegraphics[width=3.8in]{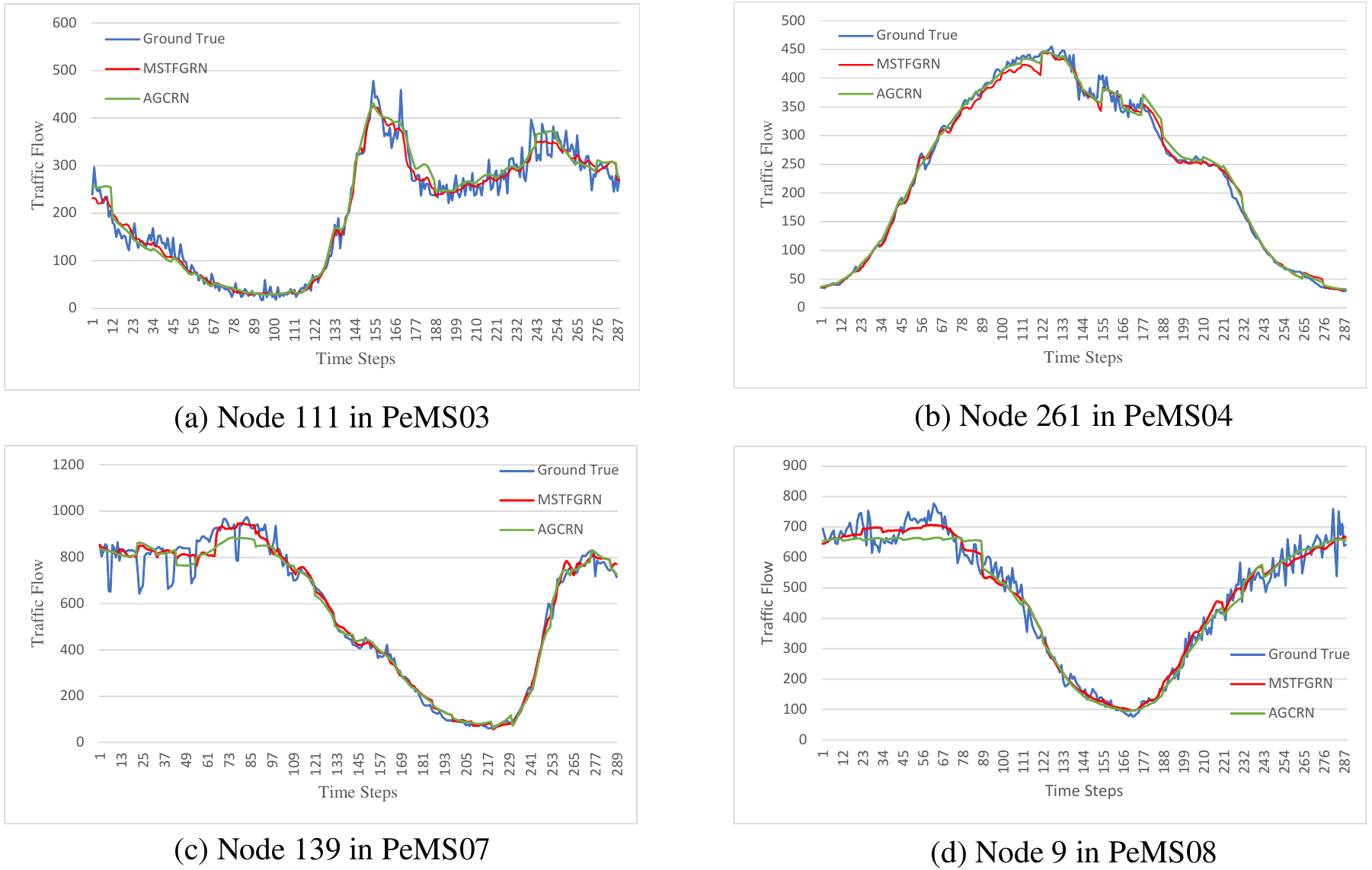}
	\caption{Traffic forecasting visualization on PeMS03, PeMS04, PeMS07, and PeMS08 dataset.}
	\label{fig_5}
\end{figure}
\end{itemize}

\subsection{Ablation Study on Model Architecture}
To further study the influence of MSTFGRN's various modules, we created four variants of the MSTFGRN-based model and compared MSTFGRN to these four variants on the PeMS04 and PeMS08 datasets. Below are the distinctions between these four model kinds. 

\begin{itemize}
	\item [1] 
	$w /o \ node \ embedding$: The model removes the node embedding operation from the SAGSAM module and uses only the predefined adjacency matrix. 
	\item [2]
	$w /o \ adjacency \ matrix$: The model removes the predefined adjacency matrix from the SAGSAM module and utilizes only the adaptive adjacency matrix.
	\item [3]
	$w /o \ reverse \ STFGRN$: The model uses only the positive STFGRN to capture Spatio-temporal correlations.
	\item [4]
	$w /o \ attention$: The model removes the global temporal attention mechanism based on MSTFGRN.
\end{itemize}
\begin{table*}[htb]
	\begin{center}
		\renewcommand{\arraystretch}{1.1}
		\caption{Results of ablation experiments on PeMS04 and PeMS08}
		\label{tab3}
		\resizebox{\textwidth}{!}{
			\begin{tabular}{cccccccc}
				\hline
				\multirow{2}{*}{Model} & Dataset & \multicolumn{3}{c}{PeMS04}  & \multicolumn{3}{c}{PeMS08}       \\ \cline{2-8}
				
				& Metrics & MAE      & RMSE     & MAPE     & MAE      & RMSE     & MAPE     \\
				\hline
				\multicolumn{2}{c}{w/o node embedding}           & 19.72    & 31.38    & 13.12\%  & 15.86    & 25.72    & 10.03\%   \\
				\hline
				\multicolumn{2}{c}{w/o adjacency matrix}          & 19.20    & 31.51    & 12.68\%  & 15.78    & 25.21    & 9.92\%  \\
				\hline
				\multicolumn{2}{c}{w/o reverse STFGRN}          & 19.52    & 31.62    & 13.35\%  & 15.66    & 25.18    & 10.00\%   \\
				\hline
				\multicolumn{2}{c}{w/o attention}          & 20.33    & 32.61    & 13.21\%  & 16.40    & 25.84    & 10.36\%  \\
				\hline
				\multicolumn{2}{c}{$\mathbf{MSTFGRN}$}       & $\mathbf{19.03}$    & $\mathbf{30.95}$    & $\mathbf{12.79\%}$  & $\mathbf{15.43}$    & $\mathbf{24.77}$    & $\mathbf{9.97\%}$   \\
				\hline
				
		\end{tabular}}
	\end{center}
\end{table*}

As shown in Table \ref{tab3}, the comparison results of the prediction performance of MSTFGRN with its four variants of the model on the PeMS04 and PeMS08 datasets are shown. We can observe that the metrics of \(w/o \ node \  embedding\) are larger than those of \(w/o \ adjacency \  matrix\), which indicates that the self-generated weighted adjacency matrix can reflect more spatial dependency information compared with the predefined adjacency matrix. However, the metrics of \(w/o \ adjacency \  matrix\) are still larger than that of MSTFGRN, which indicates that the self-generated weighted adjacency matrix can be effectively normalized using the predefined adjacency matrix. We can also observe a significant increase in the metrics of \(w/o \ reverse \  STFGRN\) compared to MSTFGRN, which indicates that capturing the contextual relevance in Spatio-temporal prediction networks is essential. In addition, we can observe that the metrics of \(w/o \  attention\) are larger. This indicates that the global temporal attention mechanism can effectively capture the global Spatio-temporal dependence and improve the model to prediction performance. Meanwhile, we compared the prediction performance of MSTFGRN with the above four model variants over various periods on the PeMS04 dataset. As shown in Figure 6, MSTFGRN achieved the best short-term prediction performance (15 MiN) and long-term prediction performance (60 Min).
\begin{figure}[H]
	\centering
	\includegraphics[width=4.6in]{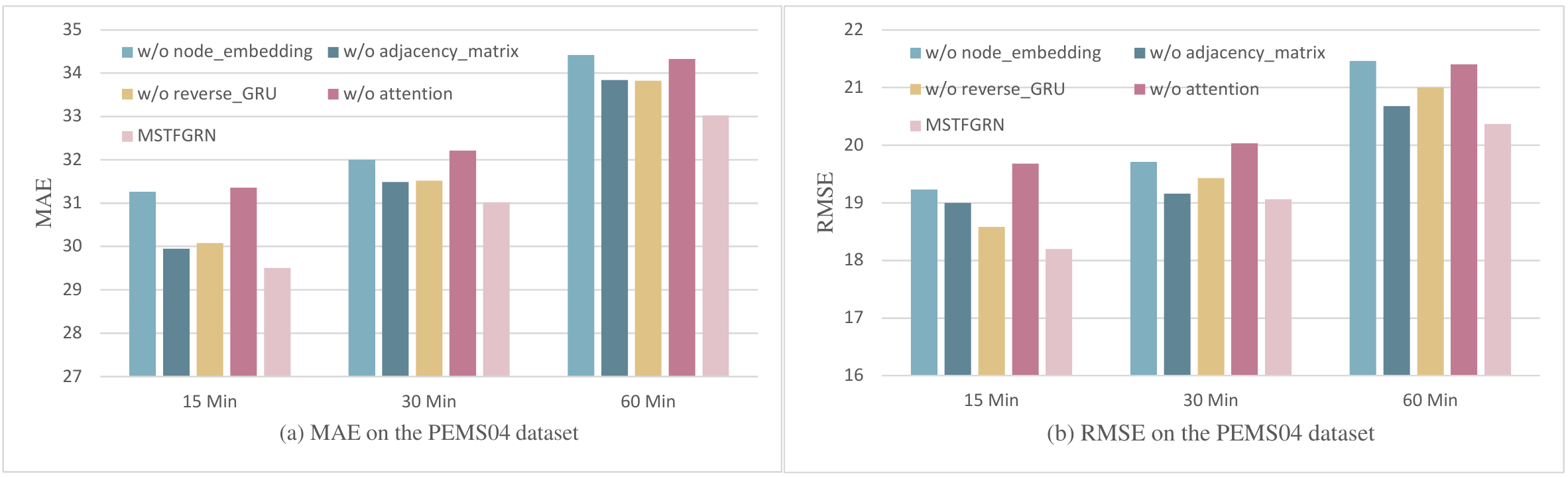}
	\caption{Ablation study on the PeMSD4 dataset.}
	\label{fig_6}
\end{figure}
Besides, the dimensionality of node embedding is an essential parameter in the SAGSAM module, which affects the quality of the spatial adjacency graph and determines whether MSTFGRN can genuinely and effectively capture the spatial correlation of the traffic road network. Figure \ref{fig_7} compares the effects of different embedding dimension numbers on the prediction performance of MSTFGRN using the PeMS04 dataset. MSTFGRN operates most efficiently with an embedding dimension of 10. When node embedding dimensions are too tiny or too high, performance degrades. This may be because when the embedding dimension is small, the information that can be contained in the node embedding module is also relatively small and cannot effectively help SAGSAM accurately derive the spatial dependence between nodes. In contrast, when the node embedding dimension is too large, the number of module parameters increases dramatically, making it impossible to optimize the model. Overall, finding the appropriate node embedding dimension is crucial to the Spatio-temporal capture capability of MSTFGRN.
\begin{figure}[H]
	\centering
	\includegraphics[width=2.9in]{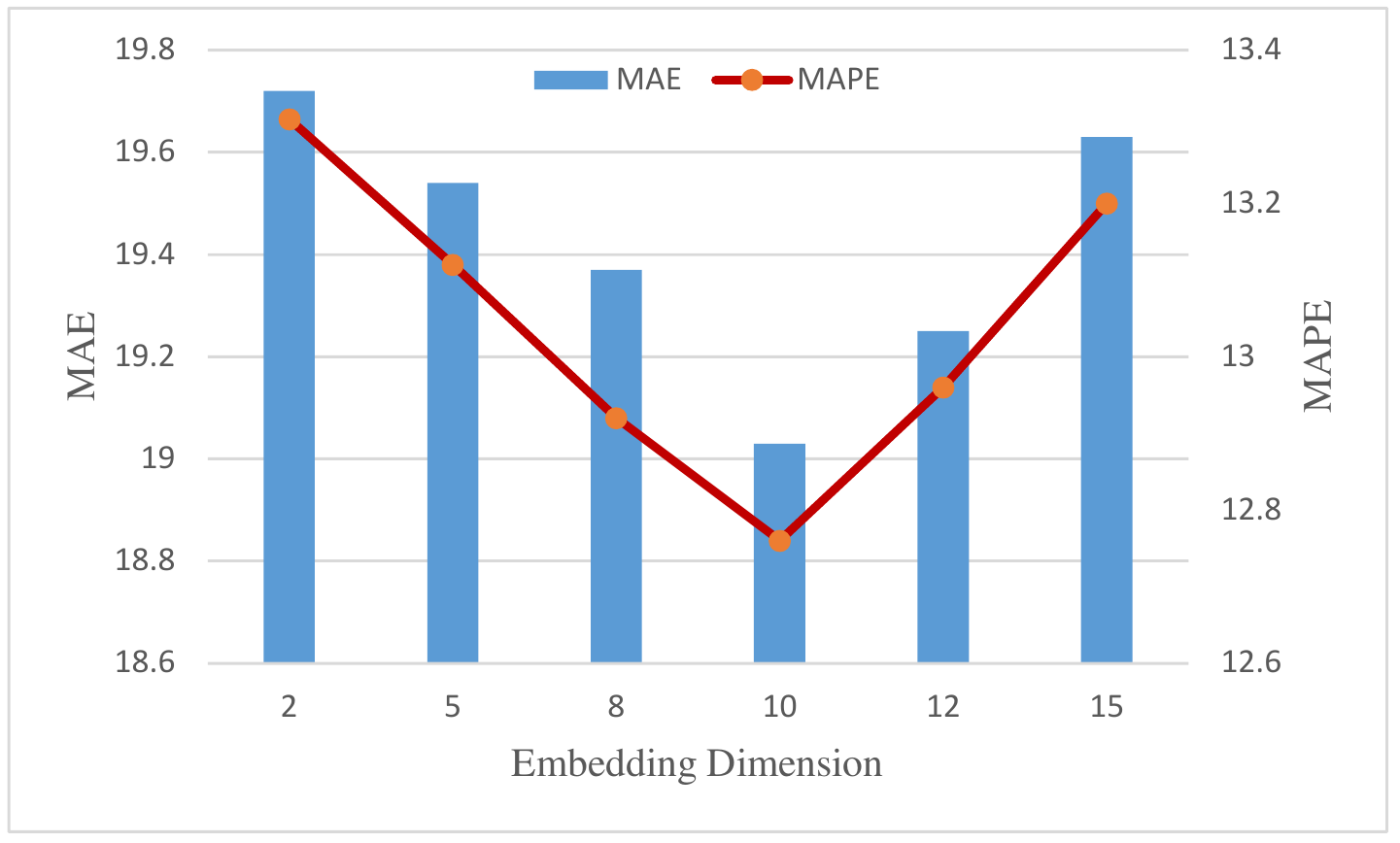}
	\caption{Impact of embedding dimensions on the PeMS04}
	\label{fig_7}
\end{figure}
\section{CONCLUSION}
This study introduces a novel paradigm for Spatio-temporal prediction using graph neural networks. The model combines an autonomously generated weighted adjacency matrix with a preset adjacency matrix and uses graph convolution techniques to capture spatial dependencies at every time step. In addition, the system offers a Spatio-temporal component to break parallel spatial dependencies on each successive time step. In conclusion, local and global Spatio-temporal dependencies are captured by imposing a global attention mechanism on each graph node. Experiments on four public transportation datasets showed that MSTFGRN produced the best prediction results overall. In our future efforts, we will concentrate on the two areas listed below: (1) Further application of the proposed framework to other Spatio-temporal prediction tasks (e.g., climate and traffic accidents); and (2) Expand the Spatio-temporal modeling capabilities of the framework for traffic prediction tasks by evaluating the incorporation of other external influences (e.g., weather, holidays, and vehicle flow) into the model to further enhance the forecast performance.

\bibliography{MSTFGRN.bib}
\end{document}